\documentclass[10pt,twocolumn,letterpaper]{article}

\usepackage{wacv}
\usepackage{times}
\usepackage{epsfig}
\usepackage{graphicx}
\usepackage{amsmath}
\usepackage{amssymb}
\usepackage{multirow}
\usepackage{threeparttable}
\usepackage{algorithm,algpseudocode}
\usepackage{autobreak}
\usepackage{authblk}
\usepackage{booktabs}

\def\wacvPaperID{681} % Enter the WACV Paper ID here

\wacvfinalcopy % *** Uncomment this line for the final submission
\ifwacvfinal
\fi

\usepackage[pagebackref=true,breaklinks=true,letterpaper=true,colorlinks,bookmarks=false]{hyperref}

\begin{document}
%%%%%%%%% TITLE
\title{Facial Emotion Recognition with Noisy Multi-task Annotations}

% \author{First Author\\
% Institution1\\
% Institution1 address\\
% {\tt\small firstauthor@i1.org}
% % For a paper whose authors are all at the same institution,
% % omit the following lines up until the closing ``}''.
% % Additional authors and addresses can be added with ``\and'',
% % just like the second author.
% % To save space, use either the email address or home page, not both
% \and
% Second Author\\
% Institution2\\
% First line of institution2 address\\
% {\tt\small secondauthor@i2.org}
% }

\newcommand*{\affaddr}[1]{#1} 
\newcommand*{\affmark}[1][*]{\textsuperscript{#1}}
\newcommand*{\email}[1]{\texttt{#1}}

\author{
Siwei Zhang\affmark[1] \quad
Zhiwu Huang\affmark[1] \quad
Danda Pani Paudel\affmark[1]  \quad
Luc Van Gool\affmark[1,2]\\
\affaddr{\affmark[1]Computer Vision Lab, ETH Z\"urich, Switzerland} \quad 
\affaddr{\affmark[2]VISICS, KU Leuven, Belgium}\\
\email{\normalsize siwei.zhang@inf.ethz.ch} \quad 
\email{\normalsize \{zhiwu.huang,paudel,vangool\}@vision.ee.ethz.ch} 
}

\maketitle
%\thispagestyle{empty}

%%%%%%%%% ABSTRACT
\begin{abstract}
Human emotions can be inferred from facial expressions. However, the annotations of facial expressions are often highly noisy in common emotion coding models, including categorical and dimensional ones. To reduce human labelling effort on multi-task labels, we introduce a new problem of facial emotion recognition with noisy multi-task annotations. For this new problem, we suggest a formulation from the point of joint distribution match view, which aims at learning more reliable correlations among raw facial images and multi-task labels, resulting in the reduction of noise influence. In our formulation, we exploit a new method to enable the emotion prediction and the joint distribution learning in a unified adversarial learning game. Evaluation throughout extensive experiments studies the real setups of the suggested new problem, as well as the clear superiority of the proposed method over the state-of-the-art competing methods on either the synthetic noisy labeled CIFAR-10 or practical noisy multi-task labeled RAF and AffectNet. The code is available at \url{https://github.com/sanweiliti/noisyFER}. 
\end{abstract}

% %%%%%%%%% BODY TEXT
\section{Introduction}
\label{sec:intro}

\begin{figure}[t]
    \centering
    \includegraphics[scale = 0.10]{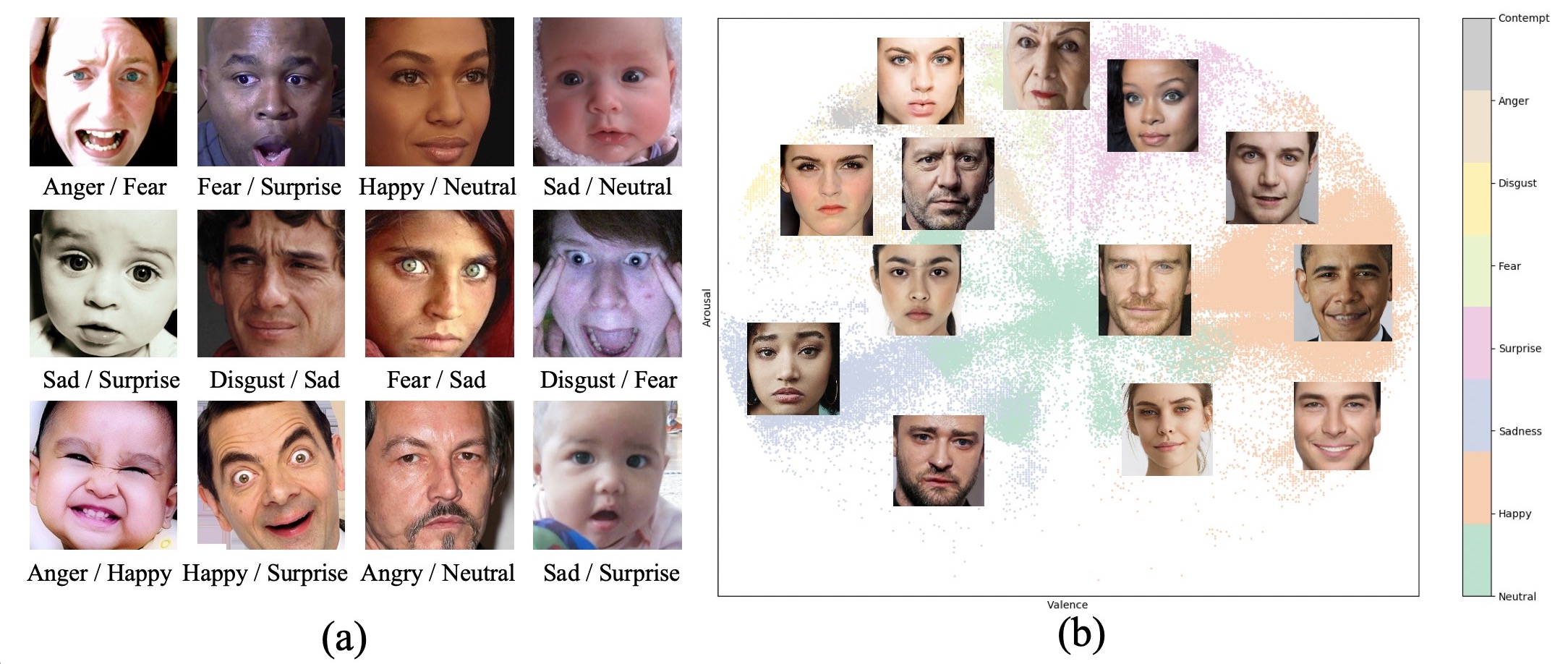}
    \caption{Illustration of (a) biased annotations on categorical emotions \cite{mollahosseini2017affectnet}, (b) the association between categorical and dimensional emotions.}
    \label{fig:fernl1}
\vspace{-0.5cm}
\end{figure}

As a window to the mind, face expresses various human emotions and intents in everyday life. This leads to a common assumption that a person's emotional state can be readily inferred from his or her facial movements. To automatically recognize facial expressions of emotions, large amounts of datasets (e.g., \cite{lucey2010extended,valstar2010induced,zhao2011facial,dhall2011static,fabian2016emotionet,mollahosseini2017affectnet,li2017reliable,kollias2018aff,kollias2019expression}) and machine learning methods (e.g., \cite{zeng2008survey,shan2009facial,pantic2009facial,chen2013optimistic,liu2014facial,jung2015joint,zhao2016peak,zeng2018facial,li2018occlusion}) have been suggested. 
However, how people communicate and understand basic categorical emotions (i.e., happiness, sadness, anger, disgust, surprise and fear) vary substantially across cultures, situations, and even across people within the same situation. In addition, it is often challenging for people to distinguish several facial emotion pairs such as, anger vs. disgust, and surprise vs. fear. Fig.~\ref{fig:fernl1}(a) shows some examples where two experts perceive emotions differently. Moreover, the challenge of correctly annotating emotions increases dramatically when people are asked to annotate dimensional emotions, i.e., valence and arousal values, which are typically defined within a continuous range of $[-1, 1]$. Therefore, biased annotations of facial expressions are inevitable and ubiquitous. On the other hand, as illustrated in Fig.\ref{fig:fernl1} (b), the categorical and dimensional labels have close correlation, despite of them being targeted for different tasks, i.e., emotion recognition vs. affect prediction.  

In this paper, we explore a new problem of facial emotion recognition from noisy multi-task labels. To reduce human efforts for labelling, we suggest to make use of cheap annotations, allowing noisy labels of the kind typically obtained from the web collection or using non-expert annotators. In fact, such labels for emotion already do exist in the form of different coding models across various datasets~\cite{fabian2016emotionet,mollahosseini2017affectnet,li2017reliable}. Among many, two most commonly used facial emotion coding models are  categorical and dimensional. Therefore,  we suggest to learn emotion recognition from multi-task labels of these two kinds. Our general observation indicates that the association between image data and the available emotion labels, although noisy, are often correct when compared to the incorrect ones, for both coding models considered. Unfortunately though, directly learning from such labels results in undesired outcome during inference.   
Inspired by this observations, we advice to benefit from multiple labels per image hereby posing a problem of learning from related multi-task labels of different encoding models. Our proposed formulation addresses this problem from the perspective of joint distribution match, which aims at exploiting the correlations among data and multi-task labels to reduce the impact of the labels' noise. 

While the formulation facilitates us to better treat the new problem, it is challenging to achieve a unified model to play the trade-off between the basic emotion prediction and the constraint on joint distribution match.
Furthermore, it is known that exact modeling of probability density function is computationally intractable \cite{salakhutdinov2009deep} for all but the most trivial cases. Additionally for our case, due to the high heterogeneity between data and multi-task labels, it is non-trivial to learn their joint distribution. To cope with the second issue, we instead reduce the data distribution modeling to generative modeling of distributions, using generative adversarial networks \cite{goodfellow2014generative}, which has proved the empirical strength as generative models of arbitrary data distributions. For the third issue, we follow the idea of canonical correlation analysis \cite{thompson2005canonical} to learn the correlation (or joint distribution) of the heterogeneous data using multi-stream projections, with which a common space is pursued reducing the heterogeneity. The exploration of joint distribution learning enables a good collaboration between the emotion prediction and joint distribution learning tasks in an unified adversarial learning game. 

In summary, this paper offers several contributions to address the practical case of facial emotion recognition. Major three of them are listed below:
\begin{itemize}
     \item We suggest a new problem of facial emotion recognition with noisy multi-task labels, which targets for readily available cheap multi-task annotations. 
     \item To address the proposed problem, we propose a generalized formulation with explicit joint and marginal distribution match among data and the heterogeneous multi-task labels.
     \item In our formulation, we introduce a new adversarial learning model to optimize the training of emotion prediction with the joint and marginal distribution based constraint, which is shown to be suitable for the newly proposed problem.
 \end{itemize}
\section{Related Work}
% \vspace{-0.1cm}
\vspace{0.2em}
\noindent\textbf{Learning from noisy labels.} 
A number of approaches have been proposed for learning noisy labels. There are several different lines to address this issue.
The first line aims to parameterize the noise by a transition matrix indicating the probability to mislabel one class into another class \cite{sukhbaatar2014training,goldberger2016training,tanno2019learning,patrini2017making}. \cite{patrini2017making} performs a forward correction and a backward correction to the noisy labels given a transition matrix. \cite{sukhbaatar2014training,goldberger2016training} add a linear layer on top of the network to learn the transition probability jointly with the prediction network. 
The second line learns the label distribution \cite{gao2017deep,tanaka2018joint,yi2019probabilistic}. For example, \cite{gao2017deep} introduces deep label distribution learning by assuming a predefined distribution for the true labels and minimizes the Kullback-Leiber divergence.
\cite{tanaka2018joint,yi2019probabilistic} optimize both the network parameters and the label predictions as the label distributions to correct noise.
Some works explore this topic with two simultaneous networks \cite{jiang2017mentornet,malach2017decoupling,han2018co,yu2019does,ma2018dimensionality,li2019learning} updated iteratively. 
There are other methods such as sample weighting strategy to measure the confidence level of each sample \cite{jiang2017mentornet,zhang2019metacleaner}, or robust loss functions \cite{wang2019symmetric,pereyra2017regularizing,zhang2018generalized} from the loss level.
Learning with multiple noisy labels has also been explored by estimating ground truth with crowdsourcing \cite{zheng2017truth} or soft labels \cite{hu2016learning}. For emotion recognition with inconsistent labels, \cite{zeng2018facial} extents the single noise learning formulation by incorporating a separate transition matrix for each annotator. \cite{tanno2019learning} proposes to add a regularization term to minimize the trace of the transition matrix for the convergence to the true label. 

To summarize, previous works generally tackle the noise label issue by transition matrix, label distribution or other classification-specific methods, therefore are incapable of dealing with multi-task involving both discrete and continuous labels, while ignoring the joint distribution between the image and labels which potentially contains richer information to benefit learning.

\vspace{0.2em}
\noindent\textbf{Generative modeling for joint distribution learning.} Following generative adversarial network \cite{goodfellow2014generative}, there are emerging a few works that explore joint distribution learning in unsupervised inference tasks or conditional generation \cite{dumoulin2016adversarially,donahue2016adversarial,jaiswal2018bidirectional,donahue2019large,pu2018jointgan,chongxuan2017triple,gan2017triangle}. For instance,  \cite{dumoulin2016adversarially,donahue2016adversarial,brock2018large} approximate the posterior distribution by matching two joint distributions from an inference network and a generation network in an unsupervised or semi-supervised manner. With a slight difference,  \cite{chongxuan2017triple} distinguishes three joint distributions including the image-label pair from data, while \cite{gan2017triangle} incorporates two discriminators to distinguish among encoder-decoder distributions and fake-true distributions. More recently, \cite{pu2018jointgan} focuses on image synthesis from both marginal distributions and conditional distributions. 
\cite{jaiswal2018bidirectional} deals with conditional image synthesis by matching joint distribution. 
While these works have made some success on joint distribution learning, they either deal with representation learning or emphasize on image synthesis, which can hardly be applied to our noisy label learning problem without special treatment to the noisy labels. Furthermore, they generally lack abilities to learn the correlations among heterogeneous labels of different tasks.

\vspace{0.2em}
\noindent\textbf{Multi-task learning for facial emotion recognition.}
Some works utilize auxiliary tasks to facilitate facial emotion recognition \cite{devries2014multi, zhao2020deep, meng2017identity, pons2018multi, ranjan2017all, ming2019dynamic, kollias2019expression, wang2019multi, hu2018deep}.
\cite{wang2019multi} explores a multi-task learning model for expression recognition and action unit detection by automatically learning the weights for either task. 
\cite{hu2018deep} predicts emotions with facial landmark detection as a side task.
Recently a large-scale human emotion dataset Aff-Wild2 \cite{kollias2019expression} is proposed with annotations of basic expression classes, valence-arousal values and action unit labels, and multi-task models are developed to facilitate learning for each individual task.
However, these works do not treat the labelling noise problem, while noise extensively exists in common facial emotion recognition labels.

\section{Problem Formulation}

Noisy-labeled facial emotion recognition aims at training a robust model on facial images merely with noisy labels. Let $D$ be the underlying truth distribution generating $(X,Y) \in \chi \times \nu$ pairs from which $n$ iid samples $(X_1,Y_1), \ldots, (X_n, Y_n)$ are drawn. After annotating these samples with a certain bias, we obtain corrupted samples $(X_1,\tilde{Y}_1), \ldots, (X_n, \tilde{Y}_n)$, and let the distribution of $(X,\tilde{Y})$ be $D_{\rho}$. The biased annotation model $P(\tilde{Y}|Y)$ is unknown to the learner. Instances are denoted by $x \in \chi \subseteq \mathbb{R}^d$, clean and noisy labels are denoted by $y, \tilde{y}$ respectively. 

A traditional way is to model the noise directly with the noisy label distribution definition $p(\tilde{y}=j|y=i)=t_{j,i}$. The probability that an input $x$ is labeled as $j$ in the noisy data can be computed using $t_{j,i}$:
\begin{equation}
\label{eq:nl1}
\begin{aligned}
p(\tilde{y}=j|x) = \sum_i^c t_{i,j}p(y=i|x),
\end{aligned}
\end{equation}
where $c$ is the total class number. In this way, we can modify a classification model using a probability matrix $T=(t_{j,i})$ that modifies its prediction to match the label distribution of the noisy data. Let $\theta$ denote parameters of the prediction model, and $\hat{p}(y|x,\theta)$ be the prediction probability of true labels by the classification model. Then the prediction of the final model is given by
\begin{equation}
\label{eq:nl2}
\hat{p}(\tilde{y}=j|x, 
\theta, T) = \sum_i^c t_{j,i}\hat{p}(y=i|x, \theta).
\end{equation}

However, modeling label noise by the transition probability matrix $T$ has several disadvantages: firstly, the transition matrix lacks constraints for the convergence to the true $T$ \cite{tanno2019learning}, which can lead optimization to a wrong direction, secondly, the assumption that annotators mislabel one specific class into another certain class in a constant probability may not always hold. Lastly, the probability matrix is a specific case to model discrete labels and fails in the regression task, while continuous noisy labels also exist in many tasks such as affect prediction. Hence, the relationship between the images and the corresponding labels should be considered for a more generalized model which is applicable in different tasks, as well as the multi-task setting. 

To address the drawback of the traditional conditional probability modeling, we suggest to remove the transition matrix $T$ and add a constraint on the joint distribution of $(X,Y)$. As noisy labels are generally outliers of the true distribution, our goal is to infer the true joint distribution to reduce impacts of outliers. Distribution-to-distribution supervision as a regularizor makes learning more robust to noise than one-to-one supervision, therefore the key idea is to match two joint distributions. If the aim is achieved, then we are ensured that all marginals as well as all conditional distributions are aligned to some extent. Here, we consider the following two joint probability distributions over two pairs of data and labels:
\begin{equation}
\label{eq:nl3}
\begin{aligned}
\hat{p}(x, y^0, y^1=i | \theta) &= \hat{p}(x)\hat{p}(y^0, y^1=i|x, \theta),\\
\hat{q}(\tilde{x}, \tilde{y}^0, \tilde{y}^1=j | \vartheta) &= \hat{q}(\tilde{y}^0, \tilde{y}^1=j)\hat{q}(\tilde{x}|\tilde{y}^0,\tilde{y}^1=j,\vartheta),
\end{aligned}
\end{equation}
where $(x,y^0, y^1)$ is a triplet of the input facial image, its predicted non-emotion latent variable, and emotion label, $(\tilde{y}^0, \tilde{y}^1, \tilde{x})$ is a triplet of the random non-emotion latent variable, the noisy emotion label and the facial image inferred from them, and $\vartheta$ is the synthesis mapping from $(\tilde{y}^0, \tilde{y}^1)$ to $\tilde{x}$. To achieve a reliable constraint, a natural way is to optimize Kullback–Leibler or Jensen–Shannon divergence for the alignment of the two joint distributions. 

In the multi-task setting for facial emotion understanding, labels of different tasks such as categorical emotion labels (discrete emotion classes), and dimensional emotion labels (continuous valence-arousal values) convey complementary information, which can be utilized for better facial emotion recognition. Still, all types of labels are noisy and this joint distribution learning framework can be easily extended to the multi-task setting. Assume each sample $X_n$ is labeled by $T$ types of noisy labels $Y_n^1, \ldots, Y_n^T$, and the goal is to learn the joint distribution of the samples and all corresponding labels $(X, Y^1, \ldots, Y^T)$, then it is desired to optimize the alignment between the following two joint distributions:
\begin{small}
\begin{equation}
\label{eq:nl4}
\begin{aligned}
\hat{p}(x, y^0, y^1, \ldots, y^T| \theta) &=\hat{p}(x) \hat{p}(y^0, y^1, \ldots, y^T|x,\theta),\\
\hat{q}(\tilde{x},\tilde{y}^0, \tilde{y}^1, \ldots, \tilde{y}^T | \vartheta) &= \hat{q}(\tilde{y}^0, \tilde{y}^1, \ldots, \tilde{y}^T) \hat{q}(\tilde{x}|\tilde{y}^0, \tilde{y}^1, \ldots, \tilde{y}^T, \vartheta),
\end{aligned}
\end{equation}
\end{small}
where $y^0, \tilde{y}^0$ are non-emotion latent variables.

\section{Proposed Method}

As it is computationally intractable to model the explicit probability density function of data distributions of real-world data \cite{salakhutdinov2009deep}, it is generally infeasible to match the two joint distributions with the exact modeling. To overcome this issue, we resort to the generative adversarial modeling methodology \cite{goodfellow2014generative}, which models a distribution with a generator and approximates the model distribution to the true distribution with a discriminator.

\begin{figure}[t!]
    \centering
    \includegraphics[scale = 0.145]{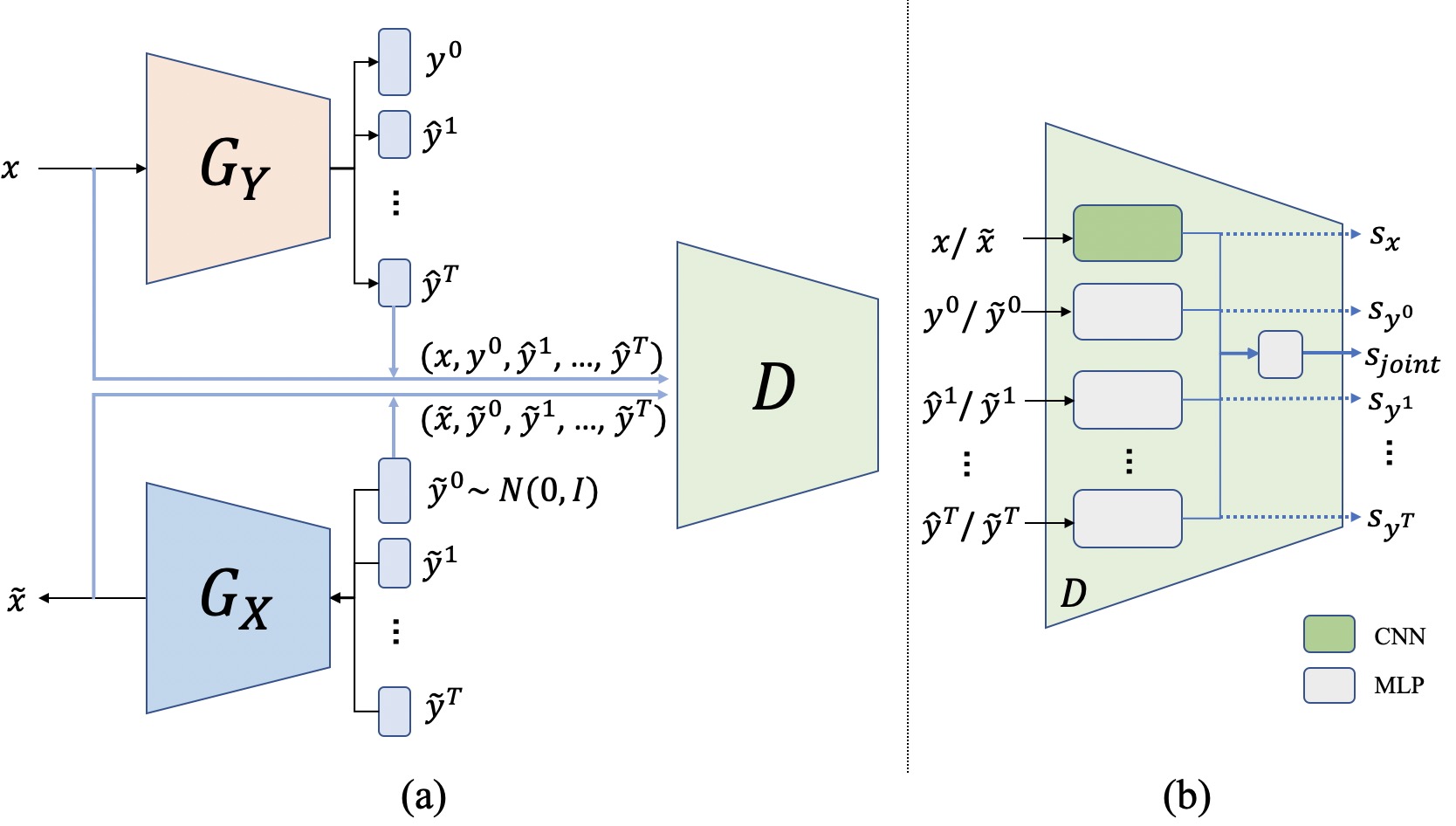}
    \caption{Illustration of (a) the overall architecture, where a pair of symmetric encoder $G_Y$ and decoder $G_X$ aims for simultaneous inference and image generation, with samples from both $G_Y$ and $G_X$ fed into $D$ to align the distributions, (b) the discriminator architecture, which is trained to learn the joint score $S_{joint}$, as well as the marginal scores $S_x, S_{y^i}$. $x$ indicates the real input image, $y^0, \hat{y}^1, \ldots, \hat{y}^T$ correspond to the predicted non-emotion and multi-task emotion labels by the encoder $G_Y$, $\tilde{y}^0$ is a Gaussian noise, $\tilde{y}^1, \ldots, \tilde{y}^T$ are the noisy labels of image $x$, and $\tilde{x}$ corresponds to the generated image by the decoder $G_X$. }
    \label{fig:method}
\vspace{-0.3cm}
\end{figure}

To model the two distributions in Eqn.~\ref{eq:nl4}, we exploit two components for the generator: One is an encoder $G_Y$ that learns the function $\theta$ to infer the clean labels from the input images, while the other is a decoder $G_X$ that learns the function $\vartheta$ to produce the facial images with the corresponding expression from the noisy labels. The architecture is illustrated in Fig.~\ref{fig:method} (a). As $G_X$ is an image generator that maps from the label space back to the image space, an additional vector $y^0$ is incorporated into our learning scheme to model other attributes of the image except for the input labels. The encoder $G_Y$ predicts the clean labels $\hat{y}^1, \ldots, \hat{y}^T$ along with the latent noise $y^0$. In the meantime, the decoder $G_X$ takes a Gaussian noise $\tilde{y}^0$, to generate image $\tilde{x}$ conditioned on the noisy labels $\tilde{y}^1, \ldots, \tilde{y}^T$:
\begin{equation}
\label{eq:generator}
\begin{aligned}
G_Y &: x \rightarrow (y^0, \hat{y}^1, \ldots, \hat{y}^T),\\
G_X &: (\tilde{y}^0, \tilde{y}^1, \ldots, \tilde{y}^T) \rightarrow \tilde{x},
\end{aligned}
\end{equation}

In order to match the joint distributions captured by the encoder and decoder, an adversarial game is played between the generator and the discriminator. In particular, the discriminator (Fig.\ref{fig:method} (b)) is designed to match the joint distribution of the group of the facial image, the noise vector, and the multi-task labels from $G_Y$ and $G_X$. For the joint distribution alignment, a natural way is to feed a pair of the group $(X, Y^0, \ldots, Y^T)$, sampled from the encoder $\hat{p}(x, y^0, \hat{y}^1, \ldots, \hat{y}^T)$ and the decoder $\hat{q}(\tilde{x}, \tilde{y}^0, \tilde{y}^1, \ldots, \tilde{y}^T)$ respectively, into the discriminator network for the adversarial training. 
Such distribution-to-distribution learning avoids overfitting to specific samples, leading to more robust inference compared to one-to-one supervision which is easily corrupted by the noisy labels.

However, the data within each group is highly heterogeneous and thus the direct concatenation on them could hurt the distribution learning. To reduce the heterogeneity among the data and multi-task labels, we suggest to exploit multiple network streams to seek their common space where their feature embeddings are less heterogeneous and hence their correlation can be better learned. As shown in Fig.~\ref{fig:method} (b), the output embeddings of all the network streams are finally fed into a common network where their joint distributions are learned by the joint score $S_{joint}$. 
Besides the advantage to leverage noisy labels of each task, the joint distribution learning also takes advantage of the implicit relationships among related tasks by $S_{joint}$ to facilitate learning for each task.
The full objective function is given by
\begin{small}
\begin{equation}
\label{eq:nl5}
\begin{aligned}
 \min_{G_X, G_Y}\max_D & ~f(G_Y(x), \tilde{y}) + \lambda(\mathbb{E}_{\hat{p}(x)}[g(D(x, G_Y(x)))] \\
 & + \mathbb{E}_{\hat{q}(y)}[h(D(G_X(\tilde{y}), \tilde{y})) ]),
\end{aligned}
\end{equation}
\end{small}
where $y=(y^0, \hat{y}^1, \ldots, \hat{y}^T)$, $\tilde{y}=(\tilde{y}^0, \tilde{y}^1, \ldots, \tilde{y}^T)$, and the integral forms of the two regularizers $\mathbb{E}_{\hat{p}(x)}[g(D(x, G_Y(x)))] $ and $\mathbb{E}_{\hat{q}(y)}[h(D(G_X(\tilde{y}), \tilde{y})) ])$ are given by
\begin{small}
\begin{equation}
\label{eq:nl5_1}
\begin{aligned}
 & \mathbb{E}_{\hat{p}(x)}[g(D(x, G_Y(x)))] \\
 & =\int\ldots\int \hat{p}(x) \hat{p}(y|x,\theta) g(D(x, y))  dxdy^0\ldots dy^T,
\end{aligned}
\end{equation}
\vspace{-2.25 mm}
\end{small}
\begin{small}
\begin{equation}
\label{eq:nl5_2}
\begin{aligned}
 & \mathbb{E}_{\hat{q}(y)}[h(D(G_X(\tilde{y}), \tilde{y}))] \\
& =\int\ldots\ \int \hat{q}(\tilde{y}) \hat{q}(\tilde{x}|\tilde{y}, \vartheta) h(D(\tilde{x},\tilde{y} ))dxd\tilde{y}^0\ldots d\tilde{y}^T.
\end{aligned}
\end{equation}
\end{small}

\begin{algorithm*} [t]
\caption{The proposed method}
\begin{algorithmic}[1]
\Require{Batch size $m$, encoder $G_Y$, decoder $G_X$, discriminator $D$, training iterations $n$, and hyperparameter $\lambda$}
\For{$i \gets 1 \textrm{ to } n$} 
    \State Sample data $(x_1, \tilde{y}_1^1, \ldots, \tilde{y}_1^T), \ldots, (x_m, \tilde{y}_m^1, \ldots, \tilde{y}_m^T)$ from the dataset
    \State sample Gaussian noise $\tilde{y}^0_1, \ldots, \tilde{y}^0_m$ from $\mathcal{N}(0,1)$
    \State $\tilde{y_j} \leftarrow (\tilde{y}^0_j, \tilde{y}^1_j, \ldots, \tilde{y}^T_j)$ for all $j$
    \State Update $G_X, G_Y$: 
    
    $\min_{G_X, G_Y} \frac{1}{m} \sum_{j = 1}^m [f(G_Y(x_j), \tilde{y_j}) + \lambda (\hat{h}(-D(x_j, G_Y(x_j))) + \hat{h}(D(G_X(\tilde{y}_j), \tilde{y}_j)))]$
    \State Update $D$:  
    
    $\max_{D} \frac{1}{m} \sum_{j = 1}^m
        [g(D(x_j, G_Y(x_j))) + h(D(G_X(\tilde{y}_j), \tilde{y_j})))]$

\EndFor
\end{algorithmic}
\label{alg}
\end{algorithm*}

The proposed generator and discriminator enable us to optimize the emotion prediction based loss and the distribution match based constraint within a unified framework. On one hand, the encoder $G_X$ is trained to predict the clean labels. On the other hand, the discriminator $D$ learns to align the distributions by distinguishing them, while the encoder $G_Y$ and decoder $G_X$ are trained jointly to fool the discriminator in an adversarial game. According to such scheme, we exploit a min-max objective function for our final solution:
\begin{equation}
% \small
\label{eq:nl5}
\begin{aligned}
 \min_{G_X, G_Y}  & f(G_Y(x), \tilde{y})  + \lambda (\mathbb{E}_{\hat{p}(x)}[\hat{h}(-D(x, G_Y(x)))]\\
 & + \mathbb{E}_{\hat{q}(y)}[\hat{h}(D(G_X(\tilde{y}), \tilde{y})) ]) \\
 \max_D  ~ & \mathbb{E}_{\hat{p}(x)}[g(D(x, G_Y(x)))] + \mathbb{E}_{\hat{q}(y)}[h(D(G_X(\tilde{y}), \tilde{y})) ], 
\end{aligned}
\end{equation}
where $y=(y^0, \hat{y}^1, \ldots, \hat{y}^T)$, $\tilde{y}=(\tilde{y}^0, \tilde{y}^1, \ldots, \tilde{y}^T)$, $\lambda$ plays a trade-off between the multi-task based prediction loss $f(G_Y(x), \tilde{y})$ and the joint distribution learning constraint with the two component functions $g(D(x, G_Y(x)))$, $h(D(G_X(\tilde{y}), \tilde{y}))$ and $\hat{h}(D(G_X(\tilde{y}), \tilde{y}))$, which correspond to adversarial losses \cite{dong2019towards}. We adopt the hinge loss for the adversarial loss that is commonly used by exiting works like \cite{miyato2018spectral,zhang2018self}.
For the multi-task learning, $f(G_Y(x), \tilde{y})$ can be realized by applying a regular cross entropy loss to the emotion recognition task, with a similarity loss for the continuous labels, which can be replaced by any target-specific loss, such as the Concordance Correlation Coefficient (CCC) loss \cite{ringeval2015avec,valstar2016avec} for affect prediction, which proves to be more efficient than the common L2 loss \cite{mollahosseini2017affectnet,kollias2018aff,kollias2019deep}.
Formally, the corresponding functions $f$, $g$, $h$, $\hat{h}$ applied for the minimization on the generator are given by 
\begin{equation}
\label{eq:nl7}
\begin{aligned}
f((y_d, y_c), (\tilde{y}_d, \tilde{y}_c)) &= L_{CE}(y_d, \tilde{y}_d) + \gamma L_{sim}(y_c, \tilde{y}_c) \\
g(z) &= \min(0, z-1), \\
h(z) &= \min(0, -z-1), \\
\hat{h}(z) &= -z, 
\end{aligned}
\end{equation}
where $y_d, \tilde{y}_d$ denote the discrete labels, $y_c, \tilde{y}_c$ indicate the continuous labels, and $z$ represents the discriminator's output. $L_{CE}$ and $L_{sim}$ denote the cross entropy loss and the similarity loss for discrete and continuous labels respectively.
$\gamma$ is the trade-off between $L_{CE}$ and $L_{sim}$.

Although aligning the joint distribution by $S_{joint}$ implicitly matches the marginal distribution, the noise and true label distributions can be very different among noisy heterogeneous labels. In this case, explicit marginal distribution alignment is beneficial. 
The alignment between the synthetic image distribution and the real image distribution can guide the decoder to generate more realistic images, meanwhile enforcing the distribution of decoder's predicted labels $\hat{y}^1, \ldots, \hat{y}^T$ to match the distribution of the noisy labels $\tilde{y}^1, \ldots, \tilde{y}^T$. 
Accordingly, each individual network stream within $D$ is expected to learn the corresponding marginal distribution by the marginal scores $\{S_x, S_{y^0}, S_{y^1}, ..., S_{y^T}\}$ as shown in Fig.~\ref{fig:method} (b).
Since we consider the facial emotion recognition as the target task, we use affect prediction as the auxiliary task to benefit the target task from both the image-to-label relationship and task-to-task relationship. 
The algorithm of the proposed method is illustrated in Alg.~\ref{alg}. 

\section{Evaluation}
% \vspace{-0.1cm}
We evaluate the proposed model in two scenarios: (1) a synthetic noisy labeled dataset (CIFAR-10 \cite{krizhevsky2009learning}) for image classification; (2) two practical facial emotion datasets (RAF \cite{li2017reliable} and AffectNet \cite{mollahosseini2017affectnet}) for facial emotion recognition. 
For a more real-world setup, we do not use clean validation labels for model selection, and thus the finally converged trained model of each comparing method is used directly for evaluation in all experiments. Please refer to supplementary material for more implementation and architecture details.

\subsection{Evaluation on Synthetic Noisy Labeled Dataset}

% \vspace{0.2em}
\noindent\textbf{Experiment setup. }
Following \cite{zeng2018facial}, the CIFAR-10 dataset \cite{krizhevsky2009learning} for image classification is selected to build the synthetic noisy labeled dataset, as a simulation case to study model behavior with multiple increasing noise. 
CIFAR-10 includes 60,000 images of size 32x32 in 10 different categories, among which 50,000 are used for training and 10,000  for the test. Images in CIFAR-10 are labeled only for image classes, from which we generate three different sets, in order to simulate our multi-task scenario. Our simulation creates three training sets with different noisy labels, by randomly flipping 20\%, 30\%, and 40\% (each for one set) of the corresponding clean labels. We do not introduce any noise in the test set. The modified noisy labels are uniformly selected across classes. Although this setup is not ideal in the sense of multi-task labels, the proposed model is still applicable, where three inconsistent noisy labels are treated as $\tilde{y}^1, \tilde{y}^2,$ and $\tilde{y}^3$, respectively. $\lambda$ is set as 0.8 in the experiment.

\noindent\textbf{Baselines. }
As the encoder $G_Y$ is a VGG-backboned network, we compare the proposed model with the following baselines: VGGNet \cite{simonyan2014very} trained on clean labels; VGGNet trained on the majority vote of the three noisy labels; VGGNet trained with all noisy labels; auxiliary image regularizer model (AIR) \cite{azadi2015auxiliary},  symmetric cross entropy loss method (SCE) \cite{wang2019symmetric}; Co-teaching method \cite{han2018co}; and  LTNet~\cite{zeng2018facial}. Among the competing methods, LTNet is proposed to deal with inconsistent labels, while other methods mainly tackle the single noisy label issue. 
We adapted the remaining methods to multiple noisy labels setting, by combining losses of all label sets.

% \vspace{0.2em}
\noindent\textbf{Results and analysis. }
The results are summarized in Tab.~\ref{table:cifar10}. Note that the results of AIR and LTNet are directly from \cite{zeng2018facial}, because AIR's result cannot be reproduced by ourselves and LTNet has not released its official code. 
As can be observed, noisy labels severely hurt the learning performance, when used without further treatment. 
The VGGNet model trained with majority voting label performs poorly because the majority voting only decreases noise ratio, but cannot model the label distribution.
The VGGNet model trained with multiple noisy labels also tends to overfit on the noisy training set by one-to-one supervision, leading to a degraded performance in the clean test set.  
AIR is not trained end-to-end and therefore difficult to optimize. SCE and Co-teaching deal with single noise label issue, from the perspective of robust loss function or complementary networks, hence they lack generalization abilities to the multi-label or multi-task setting. 
LTNet, which is specifically designed for inconsistent label setting, performs comparable to the VGGNet model trained with clean labels. However, it is not applicable for continuous labels as it models noise by the transition matrix. 
In comparison, the proposed model trained with multiple noisy labels achieves the best result among compared methods, and is comparable to the model trained using clean labels. Note that our method is not restricted by the number or task types or the noisy labels. 

\begin{table}[t]
\begin{center}
\caption{Test accuracy on CIFAR-10 synthetic dataset.}
\label{table:cifar10}
\begin{tabular}{l|l|c}
\toprule[1pt]
Training data & Model & Test acc (\%)\\
\midrule[1pt]

Clean labels& VGGNet \cite{simonyan2014very} & 88.55\\ 
\midrule

\multirow{7}*{Multi noisy labels} 
& VGGNet-major vote & 82.36\\
& VGGNet & 80.23 \\
& AIR \cite{azadi2015auxiliary} & 76.37 \\ 
% & NAL \cite{} & 84.4 \\
& SCE \cite{wang2019symmetric} & 86.34  \\
& Co-teaching \cite{han2018co} & 84.21 \\ 
& LTNet \cite{zeng2018facial} & 87.23 \\ 
& Proposed & \textbf{87.90} \\ 
\bottomrule[1pt]
\end{tabular}
\end{center}
\vspace{-0.8cm}
\end{table}
% \vspace{-0.1cm}
% \setlength{\tabcolsep}{1.4pt}

\begin{figure}[t]
    \centering
    \includegraphics[scale = 0.5]{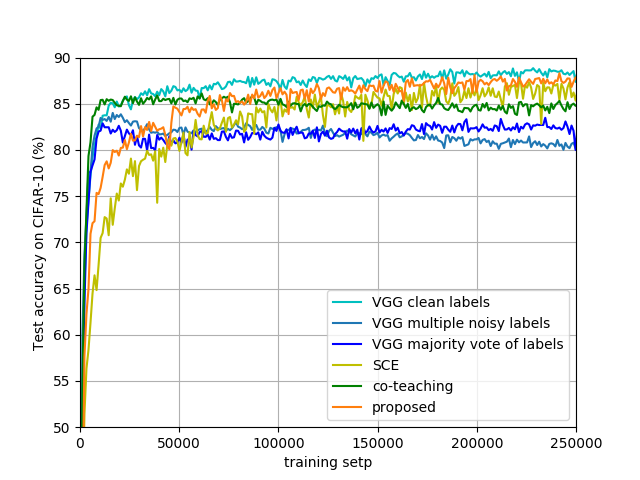}
    \caption{Test accuracy vs. training steps on CIFAR-10 synthetic noisy dataset.}
    \label{fig:test_curve}
\vspace{-0.5cm}
\end{figure}

The test accuracy curves of the baselines and the proposed model over training steps are visualized in Fig.~\ref{fig:test_curve}. Note that the baseline models trained with noisy labels first rise to a peak accuracy quickly and decrease later. In contrast, the test accuracy of the proposed model continues increase over the training steps. Furthermore, we observed that the one-to-one cross entropy loss of the proposed model does not converge to zero, instead settles at a relatively high value. However, the cross entropy loss of VGGNet baseline models converge to almost zero, which indicates overfitting. This observation demonstrates that the joint distribution learning proposed in this paper can avoid the negative influence of the noisy labels.

\subsection{Evaluation on Facial Emotion Dataset}
The distribution of the synthetic noise in CIFAR-10 dataset is straightforward and only includes one task, while labelling noise can have various patterns in practical emotion datasets due to many reasons (different annotators, labelling protocols, challenges in distinguishing certain expressions, etc.).
To study the effectiveness of the proposed model in the emotion recognition task with noisy multi-task labels, we evaluate the model in real-world settings.

\noindent\textbf{Experiment setup. }
A cross-dataset evaluation protocol is proposed by LTNet~\cite{zeng2018facial} to train the model on the combination of AffectNet~\cite{mollahosseini2017affectnet}, RAF~\cite{li2017reliable} training set (and some unlabeled facial images), and test on several emotion recognition dataset test sets including AffectNet and RAF. However, such setting is not suitable to our new problem: this training setting assumes that the labelling of AffectNet and RAF are biased and inaccurate, while reporting test accuracy on both AffectNet validation set and RAF test set to demonstrate the superiority of the proposed framework assumes that the same labelling on these two datasets can result in clean labels for test, which is contradictory with the training. Therefore, for the suggested new problem, we propose a more appropriate evaluation procedure on RAF and AffectNet. We use machine labeling by pretrained models, which inherently introduces noise due to domain gap and human-machine disagreements.

\begin{table*}[t]
\begin{center}
\caption{Evaluation results on facial emotion datasets. 
Single-task refers to models trained only with categorical expression labels or valence-arousal labels, and multi-task refers to models trained with both expression and valence-arousal labels. 
Emotion/Acc (\%) denotes test accuracy of categorical expression prediction, the higher the better. 
VA/CCC, VA/MSE denote CCC and MSE metrics of valence-arousal prediction respectively, the higher the better for CCC and the lower the better for MSE.
(\textbf{Bold}: best, \underline{Underline}: second best) }
\label{table:emotion experiment}
\begin{tabular}{l|l|c|ccc}
\toprule[1pt]
Setting & Model & RAF-base  &  \multicolumn{3}{c}{AffectNet-base} \\
\midrule
Task/Metric &  & Expression/Acc (\%) & Expression/Acc (\%) & VA/CCC & VA/MSE \\

\midrule[1pt]
\multirow{4}*{Single-task} 
& VGGNet \cite{simonyan2014very} & 72.64 &  43.42 & 0.6254 & 0.1438\\
& SCE \cite{wang2019symmetric} & 73.96  &  42.87  & - & -\\
& Co-teaching \cite{han2018co} & \underline{75.43}  &   42.36 & - & -\\
& Proposed & 74.02   &  \underline{44.52} & \underline{0.6354} & \underline{0.1284}\\ 

\midrule
\multirow{2}*{Multi-task} 
& VGGNet & 73.15 &  43.36 & 0.6263 & 0.1354\\
& Proposed & \textbf{76.10} & \textbf{46.08}  & \textbf{0.6727} & \textbf{0.1248}\\
\bottomrule[1pt]
\end{tabular}
\end{center}
\end{table*}

% \vspace{0.2em}
\noindent\textbf{Baselines. }
As existing methods for noisy label learning either are not applicable, or cannot be easily adapted to the multi-task setting or continuous labels, we choose to train two of state-of-the-art methods for single discrete noisy label learning, i.e., SCE \cite{wang2019symmetric}, Co-teaching \cite{han2018co}, only with discrete expression labels. For comparison, we also train the VGGNet model and a degraded version of our model in the single-task setup only with discrete expression labels or continuous valence-arousal labels. 
To evaluate the proposed multi-task model's full effectiveness, we train it with both expression and valence-arousal labels. With no existing work applicable for multi-task noisy label learning with both discrete and continuous labels, we train another VGGNet model with multi-task labels with loss combination of both tasks for comparison.

The \textbf{RAF} dataset consists of 15,339 real-world facial images collected from Flickr, including 12,271 images for training and 3,068 images for testing. Each image was labelled as one of the seven basic facial emotions (i.e., neutral, happy, sad, surprise, angry, fear, disgust) by about 40 independent annotators, followed by an EM algorithm to assess the reliability of each annotator. The \textbf{AffectNet} dataset is a multi-task in-the-wild dataset including around 450,000 training images, and 5,500 validation images. Each image in AffectNet is labeled by two labels: one of the eight discrete expression classes (seven basic expressions with an additional contempt class), and the continuous valence-arousal values. The labels for the test set is not public therefore we only report accuracy results on the validation set, which is not seen during training. We select images with the seven basic expression labels in the experiment, resulting in 283,910 training images and 3,500 test images. 

We create two scenarios for evaluation: (1) the RAF training set is relabeled for both expression classes and valence-arousal values by an AffectNet pretrained model for training, and we keep RAF test set unchanged for evaluation; (2) the expression classes of AffectNet training set is relabeled by a RAF pretrained model for training, together with the original valence-arousal labels of AffectNet, and the trained model is evaluated on the AffectNet validation set. We denote the two experiment scenarios by \textbf{RAF-base} and \textbf{AffectNet-base} respectively. 
For valence-arousal prediction, we only report results in AffectNet-base case since no human valence-arousal labels are available on RAF.

\begin{table*}[t]
\begin{center}
\caption{Ablation study for different components of the proposed model: w/o joint, w/o marginal and w/o $G_X$ respectively remove $S_{joint}$, $\{S_x, S_{y^0}, S_{y^1}, ...\}$ and $G_X$ in Fig. \ref{fig:method}.}
\label{table:ablation study}
\begin{threeparttable}
\begin{tabular}{l|c|c|c|c|c}
\toprule[1pt]
 Setting & VGGNet & w/o joint & w/o marginal & w/o $G_X$ & Proposed \\
\midrule[1pt]
Single-label CIFAR-10\tnote{*} & 77.82 & 77.93 & 83.65 &  83.51 & \textbf{84.78} \\
Multi-label CIFAR-10 & 80.23 & 79.10 & 85.48 & 87.88 & \textbf{87.90} \\ 
Single-task RAF-base & 72.64 & 72.28 & 73.19 & 73.37 & \textbf{74.02} \\ 
Multi-task RAF-base & 73.15 & 73.24 & 74.40 & 74.46 & \textbf{76.10} \\ 
\bottomrule[1pt]
\end{tabular}
\begin{tablenotes}
        \footnotesize
        \item[*] Use one single noisy label set with 20\% noise for training 
      \end{tablenotes}
 \end{threeparttable}
\end{center}
\vspace{-0.3cm}
\end{table*}

\begin{figure*}[ht!]
    \centering
    \includegraphics[scale = 0.54]{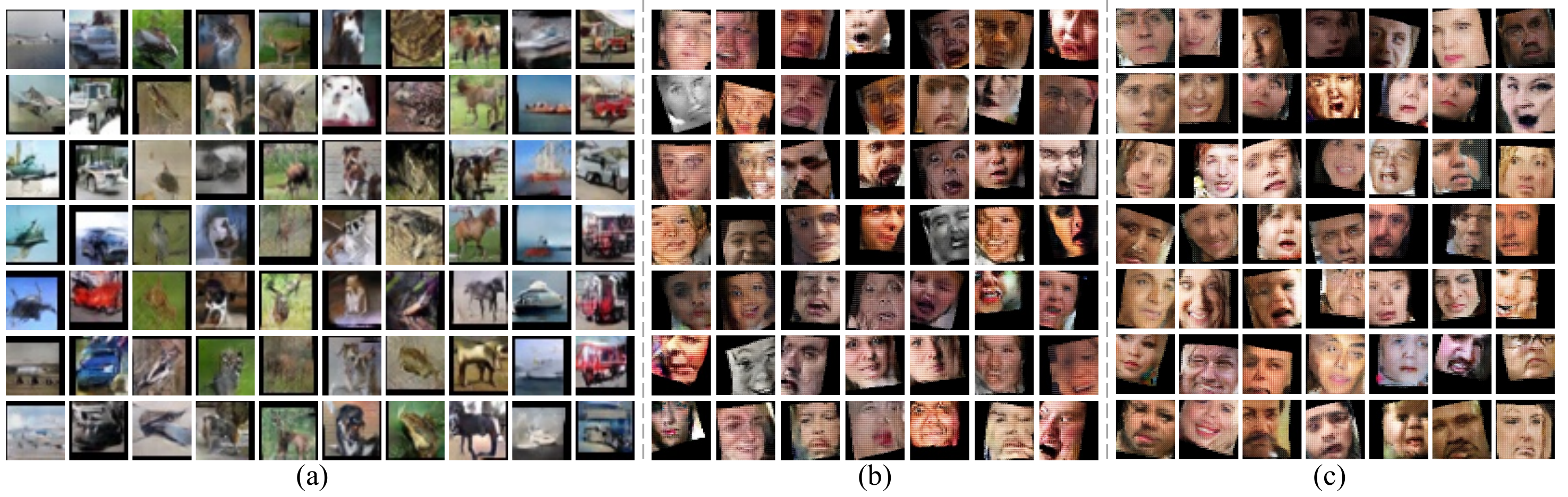}
    \caption{Samples synthesized by the decoder $G_X$ conditioned on the input label: (a) samples from CIFAR-10, and each column conditioned on input class labels as \textit{airplane, automobile, bird, cat, deer, dog, frog, horse, ship, truck} from left to right; (b)(c) samples from RAF/AffectNet, respectively, and each column conditioned on input expression labels as \textit{neutral, happy, sad, surprise, fear, disgust, anger} from left to right. The rotations and boundaries are due to data augmentation during training.}
    \label{fig:gen_imgs}
% \vspace{-0.2cm}
\end{figure*}

% \vspace{0.2em}
\noindent\textbf{Results and analysis. }
The experiment results on facial emotion datasets are illustrated in Tab.~\ref{table:emotion experiment}. 
For the discrete expression prediction, in the single-task RAF-base case, SCE and the proposed model achieve similar performance, all higher than VGGNet, and Co-teaching achieves the best performance. Co-teaching trains two networks simultaneously, with each one selecting small-loss instances (more likely to have correct labels) and teaching its peer network. However, such instances are less likely to occur with increasing noise intensity. Therefore, Co-teaching is a strong baseline for the single-task RAF-base case, where the noise level is relatively low. 
For the AffectNet-base case where the model is exposed to higher noise, SCE and Co-teaching face difficulties for optimization with more complicated noise distributions, and perform even worse than VGGNet. Moreover, it is non-trivial to adapt these two baselines to multi-task. In contrast, the proposed model takes full advantage from multi-task noisy labels and achieves the best performance.
For the continuous valence-arousal prediction, we utilize the commonly used Concordance Correlation Coefficient (CCC) and Mean Square Error (MSE) as the evaluation metrics. We can observe that the proposed model trained in the multi-task setting significantly improves prediction performance compared with the VGGNet baseline method and the single-task models.

For the case of clean labels, it is natural to combine loss functions of each task in multi-task learning \cite{kollias2019expression}. Nevertheless such simple loss combination is not necessarily beneficial in the presence of noisy labels for each task, or even harmful, which could be observed in Tab.~\ref{table:emotion experiment}. 
However, the proposed multi-task model demonstrates superior performance compared with multi-task VGGNet  and single-task models. Our joint distribution learning scheme not only leverages noise for each task individually, but also learns a robust correlation between different tasks as well as the image, to take advantage of labels of both tasks and utilize richer information for emotion recognition without being corrupted by the noise in labels of each task.

\subsection{Ablation Study}
Here we conduct the ablation study to investigate the contributions of marginal scores $\{S_x, S_{y^0}, S_{y^1}, ...\}$ and joint scores $S_{joint}$ in $D$, the decoder $G_X$ and extra labels of the proposed model. The ablation study is implemented on the following four cases: (1) single-label model on the synthetic noisy CIFAR-10 dataset with 20\% noise in the training labels, (2) multi-label model on the synthetic noisy CIFAR-10 dataset, (3) the single-task model on RAF-base case, and (4) the multi-task model on RAF-base case). Results are shown in Tab.~\ref{table:ablation study}. 
With respect to design of $D$, only marginal distribution matching (by $\{S_x, S_{y^0}, S_{y^1}, ...\}$) without joint distribution learning has marginal benefit (w/o joint), while the joint distribution learning by $S_{joint}$ in $D$ is essential to combat the noise labels (w/o marginal).
Additional marginal scores on top of $S_{joint}$ brings extra improvements by explicit marginal distribution matching (proposed). 
The decoder $G_X$ is beneficial by facilitating the joint distribution learning in a more balanced way. Here the non-emotion variable $y^0$ in Fig.~\ref{fig:method} is necessary to encode other image attributes so that $G_X$ has sufficient information for generation. 
Training with multiple labels further enhances the model by capturing the correlation among complementary labels, which is validated both in the CIFAR-10 case and practical emotion learning scenario.

\subsection{Conditional Image Synthesis}
Our proposed joint distribution learning framework learns inference and conditional image synthesis simultaneously. Since the distribution learning only serves as a regularizar in our algorithm, the generated images are not necessarily of the highest quality, but should carry explicit semantic meaning.  
Fig.~\ref{fig:gen_imgs} presents generated samples on CIFAR-10 (Fig.~\ref{fig:gen_imgs} (a)), RAF (Fig.~\ref{fig:gen_imgs} (b)) and AffectNet (Fig.~\ref{fig:gen_imgs} (c)). The decoder $G_X$ can generate correct images given the conditional label, which demonstrates that the two joint distributions could be aligned with the adversarial training.

\section{Conclusion and future work}

This paper introduces an interesting problem of facial emotion recognition with noisy multi-task annotations, which has a high potential to reduce human labelling efforts for multi-task learning. To better treat the problem, we introduce a new formulation from the view of joint distribution match. Following the suggested formulation, we exploit a new adversarial learning method to jointly optimize the emotion prediction and the joint distribution learning. Finally we study the setup of synthetic noisy labeled dataset and practical noisy multi-task datasets, and experiments demonstrate the clear advantage of the proposed method for the new problem. 
While we can roughly setup the trade-off between the emotion prediction loss and the joint distribution match based constraint, automatically adapting the balance would be interesting to study in our future work.
{\small
\paragraph{Acknowledgements.}
This work was supported by the ETH Z\"urich Fund (OK), an Amazon AWS grant, and an Nvidia GPU grant. We thank Janis Postels and Thomas Probst for their valuable discussions on this work.
}

{\small
\bibliographystyle{ieee_fullname}
\bibliography{egbib}
}

\begingroup
\onecolumn 

\appendix
\begin{center}
\Large{\bf Facial Emotion Recognition with Noisy Multi-task Annotations \\ **Supplementary Material**}
\end{center}

\setcounter{page}{1}
\setcounter{table}{0}
\setcounter{figure}{0}
\renewcommand{\thetable}{S\arabic{table}}
\renewcommand\thefigure{S\arabic{figure}}

% In the supplementary material, we present our designed network architecture details, more ablation study, confusion matrices and visualized results, as well as more study on the joint distribution learning weight.

% \section{Implementation Details}

\section{Architecture and Implementation Details}
The detailed architecture is illustrated in Fig.~\ref{fig:archi_1}. The encoder $G_Y$ is a modified VGGNet \cite{simonyan2014very} which predicts clean labels $\hat{y}^1, \ldots, \hat{y}^T$, a mean $\mu$, a variance $\sigma$, from which the latent noise vector $y^0$ is sampled as $y^0 \sim \mathcal{N}(\mu, {\sigma}^2\boldsymbol{I})$ (Fig.~\ref{fig:archi_1} (a)). 
The decoder $G_X$ (Fig.~\ref{fig:archi_1} (b)) takes the concatenation of a Gaussion noise $\tilde{y}^0 \sim \mathcal{N}(0, \boldsymbol{I})$, and all the input noise labels $\tilde{y}^1, \ldots, \tilde{y}^T$ as the input, which is first fed through a linear layer, and then upsampled by deconvolutional blocks to produce the image $\tilde{x}$. 

The discriminator $D$ consists of separate streams for each input variable, which is a CNN stream for the input image, and different multilayer perceptron (MLP) streams for the input labels or noise. The marginal scores $S_x, S_{y^0}, \ldots, S_{y^T}$ are computed as a linear transformation of the output features of each stream. In the meantime, the output features of each stream are concatenated into an MLP to produce the joint score $S_{joint}$ for the joint distribution matching. 
The CNN stream for the image includes several residual blocks and one attention block. Each residual block is a simple residual convolutional block which contains two [convolution, ReLU] blocks and one pooling operation. The attention block is a self-attention CNN block \cite{zhang2018self} which aims to utilize features from all locations, for modeling long range and multi-level dependencies. The MLP stream is an MLP block which gives the summation of outputs of four separate MLP sub-blocks.
See the detailed architecture layouts of residual block, the attention block, and MLP components of $D$ in Fig.~\ref{fig:archi_2}. 

The proposed model is implemented with PyTorch, using ADAM \cite{kingma2014adam} as the optimizer ($\beta_1=0.9$, $\beta_2=0.999$). The learning rate are set to 1e-4. $\gamma$ is always set to 1. In each iteration of the alternating optimization schedule, $G_Y$ and $G_X$ are jointly updated once, followed by two consecutive updates of $D$. See analysis of the joint distribution learning weight $\lambda$ in Section \ref{lambda study}.

\begin{figure}[h]
    \centering
    \includegraphics[scale = 0.2]{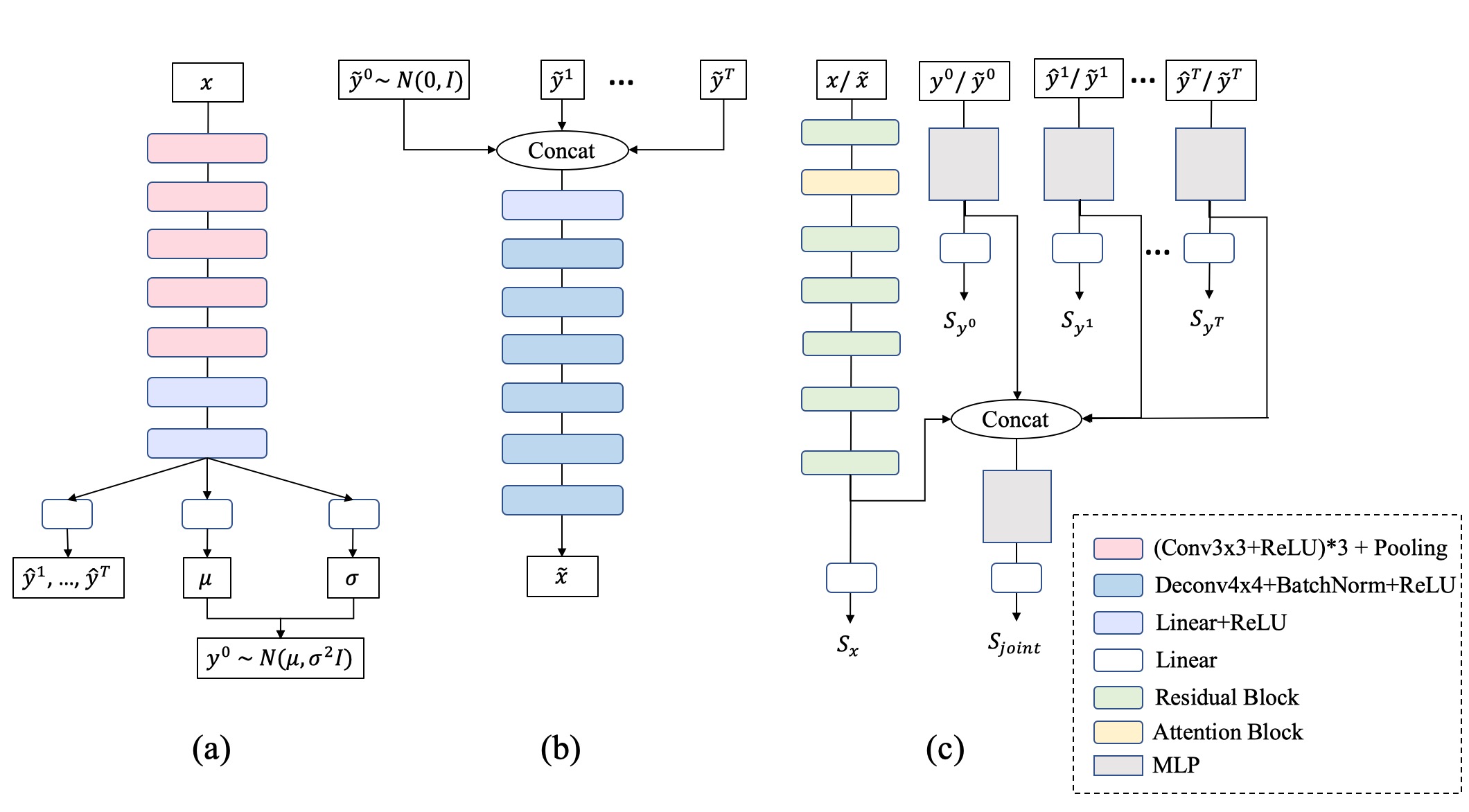}
    % \vspace{-0.3cm}
    \caption{Architecture details of (a) the encoder $G_Y$, (b) the decoder $G_X$, (c) the discriminator $D$.}
    \label{fig:archi_1}
% \vspace{-0.3cm}
\end{figure}

\begin{figure}[h]
    \centering
    \includegraphics[scale = 0.2]{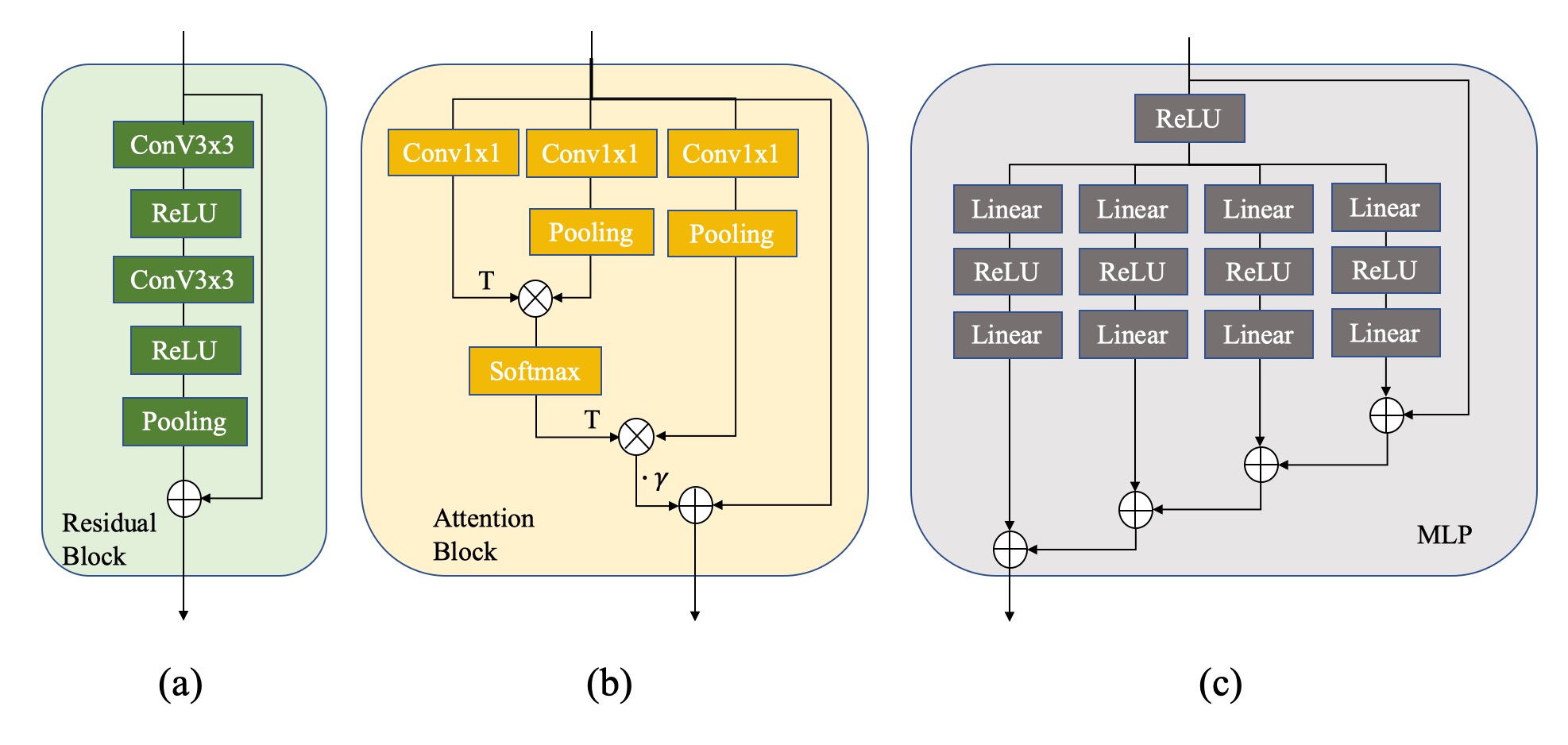}
    \vspace{-0.3cm}
    \caption{Detailed architecture layouts of components of discriminator $D$: (a) the residual block, (b) the attention block (T denotes transpose and $\gamma$ is a learnable parameter), (c) the MLP block.}
    \label{fig:archi_2}
\vspace{-0.3cm}
\end{figure}

\begin{figure}[t]
    \centering
    \includegraphics[scale = 0.2]{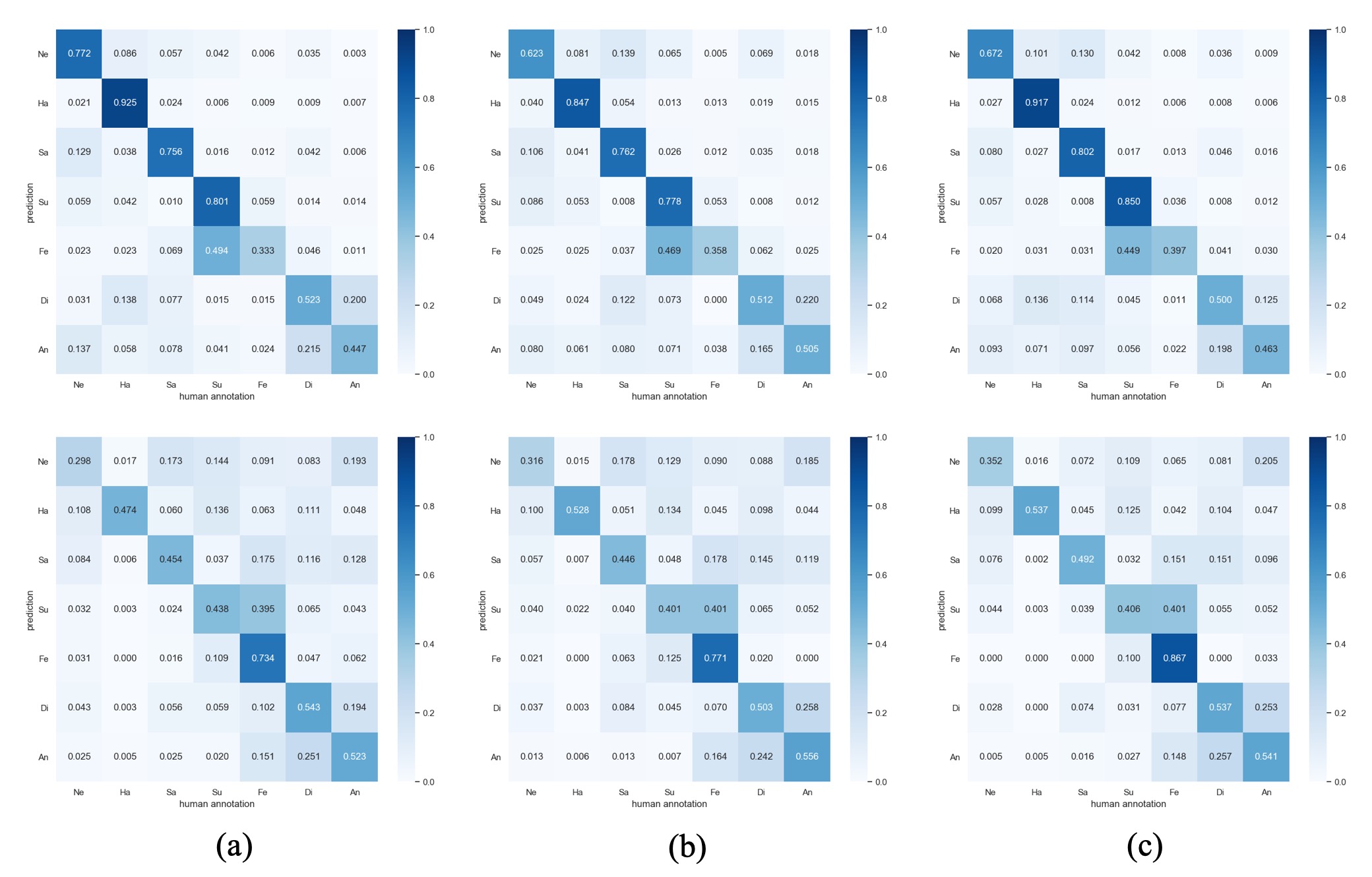}
    \caption{Confusion matrices for the RAF-base (the first row) and AffectNet-base (the second row) cases: (a) the pretrained model to create noisy training set, (b) multi-task VGGNet, (c) the proposed multi-task model trained on the noisy labels.}
    \label{fig:confusion_matrix}
% \vspace{-0.3cm}
\end{figure}

\begin{figure}[h]
    \centering
    \includegraphics[scale = 0.2]{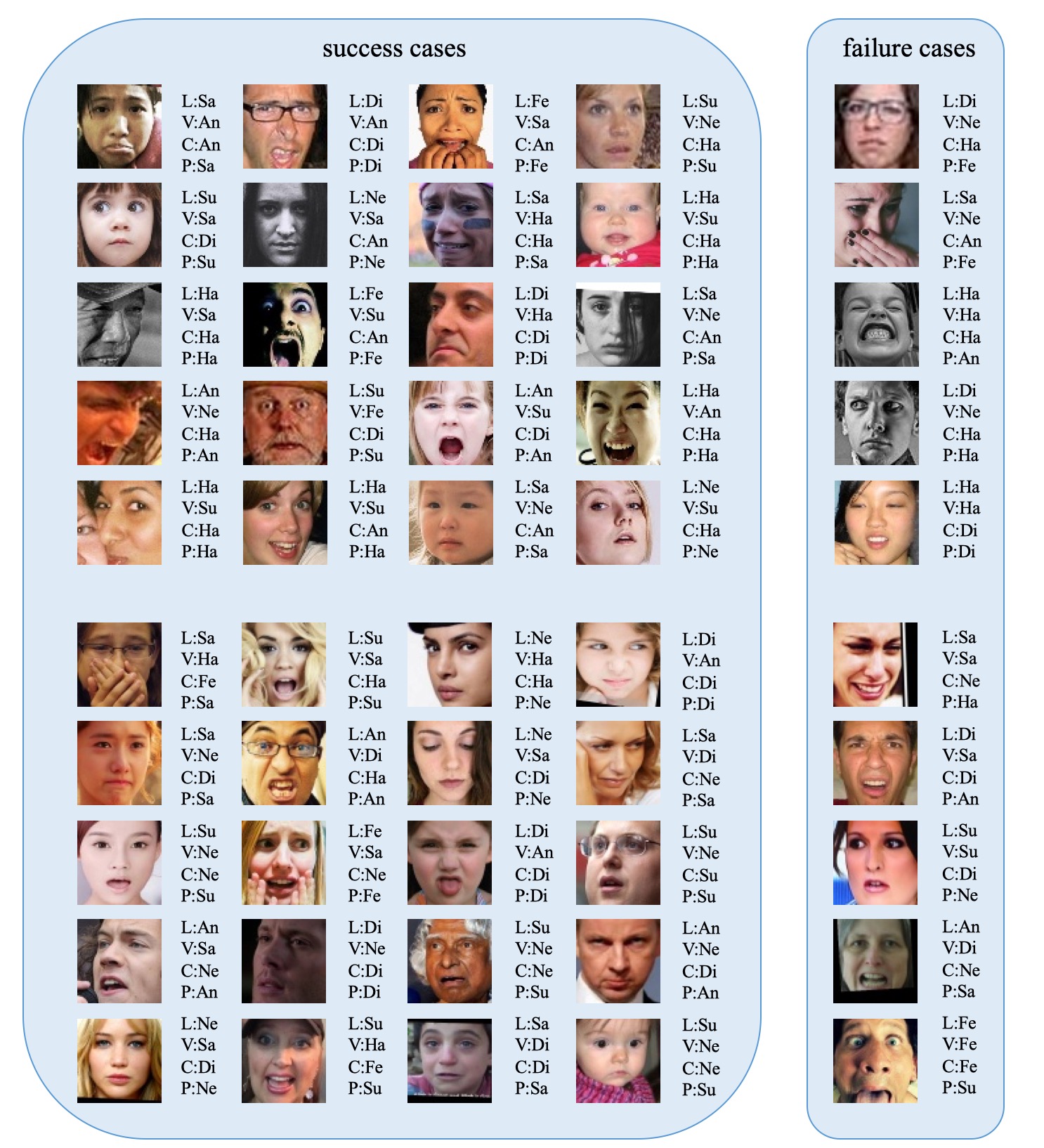}
    \caption{Examples of images and corresponding predictions on RAF-base (the first five rows) and AffectNet-base (the last five rows) cases, with the first four columns where the predictions of our proposed multi-task model are consistent with the human label in the test set, and the last column contains some failure cases of the proposed model. L: human label of test set, V: multi-task VGGNet, C: Co-teaching model \cite{han2018co} trained with only emotion class labels, P: proposed multi-task model.}
    \label{fig:selected_img}
% \vspace{-0.3cm}
\end{figure}

\section{Confusion Matrix and Visualization}
The confusion matrices for the multi-task models on facial emotion datasets in both RAF-base and AffectNet-base cases are shown in Fig.~\ref{fig:confusion_matrix}. The confusion matrices of the pretrained models which are utilized to create noisy training set are only for reference. Comparing the confusion matrices of the multi-task VGGNet models and the proposed multi-task models, the diagonal values of the proposed models are generally higher than the VGGNet models, showing that the proposed model can achieve more precise predictions by the joint distribution learning from multi-label models. However, the confusion matrices of the three different kinds of models still show similar patterns, suggesting that learning from noisy labels in practical emotion dataset is still a difficult task.

Images in Fig.~\ref{fig:selected_img} are examples of success and failure cases of the proposed multi-task model in RAF-base and AffectNet-base cases. Our proposed model gives correct labels in many cases where the Co-teaching method \cite{han2018co} or multi-task VGGNet gives inconsistent labels with the human annotation, which we believe to be relatively clean in the test set. There are also some failure cases of the proposed model, for example, where sad faces with mouth open are predicted to be happy, or happy faces with extreme activated expressions are predicted to be angry.

\section{Study on Joint Distribution Weight $\lambda$}
\label{lambda study}
Fig. \ref{fig:lambda_study} illustrates the test accuracy curve of the proposed multi-task model on both RAF-base and AffectNet-base cases with different joint distribution learning weights $\lambda$ varying among 0.2, 0.4, 0.6, 0.8 and 1.0. Since the joint distribution learning serves as a regularizar in the proposed model, the optimal choice of $\lambda$ varies among different datasets depending on the noise intensity.  
In spite of this, the value of $\lambda$ can be roughly set according to the prior knowledge about the noise intensity. For example, we are given the prior knowledge that the human annotations on AffectNet are relatively worse than those on RAF, as RAF adopts an EM algorithm to achieve a better annotation on about 40 annotations on each sample while AffectNet's annotation is merely from one single annotator. Accordingly, the pretrained model on AffectNet is assumed to offer relatively noisy labels than the one trained on RAF. Using this prior knowledge, we roughly set $\lambda$ to two values, i.e., 0.4 and 1.0, for the model training on AffectNet and RAF respectively.
In addition, a general pattern can be observed: the model performance will increase with $\lambda$ increasing in a certain range, and then decrease with $\lambda$ increasing further as the joint distribution learning overweights the task-specific losses (i.e., cross-entropy loss and CCC loss in our case).

In addition to using the prior knowledge on the noise intensity, we find the training curve of the joint distribution loss can also help us to infer a suitable $\lambda$ for better training of the proposed model.
We empirically observe that the optimal models in both RAF-base ($\lambda=1.0$) and AffectNet-base ($\lambda=0.4$) cases consistently converge to a value around 2.6 for the generator joint distribution loss, and a value around 0.45 for the discriminator joint distribution loss (see Fig.~\ref{fig:training_loss}). 
The RAF-base case is more robust when $\lambda$ is varied in a certain range, because it requires a higher $\lambda$ due to the higher noise intensity in labels.
Therefore, this observation can be used as the second strategy to roughly estimate an appropriate $\lambda$ to train the proposed model on new data with noisy multi-task labels.

If the incorrect labels form a specific distribution with a high noise ratio, the joint distribution learning might fail due to the large gap between the noise label distribution and the true label distribution. 
However, in most practical cases of facial emotion machine labels, the incorrect labels are outliers of the true distribution, and the proposed distribution-to-distribution supervision is based on such assumption, therefore robust to the noise.

\begin{figure}[h]
    \centering
    \includegraphics[scale = 0.18]{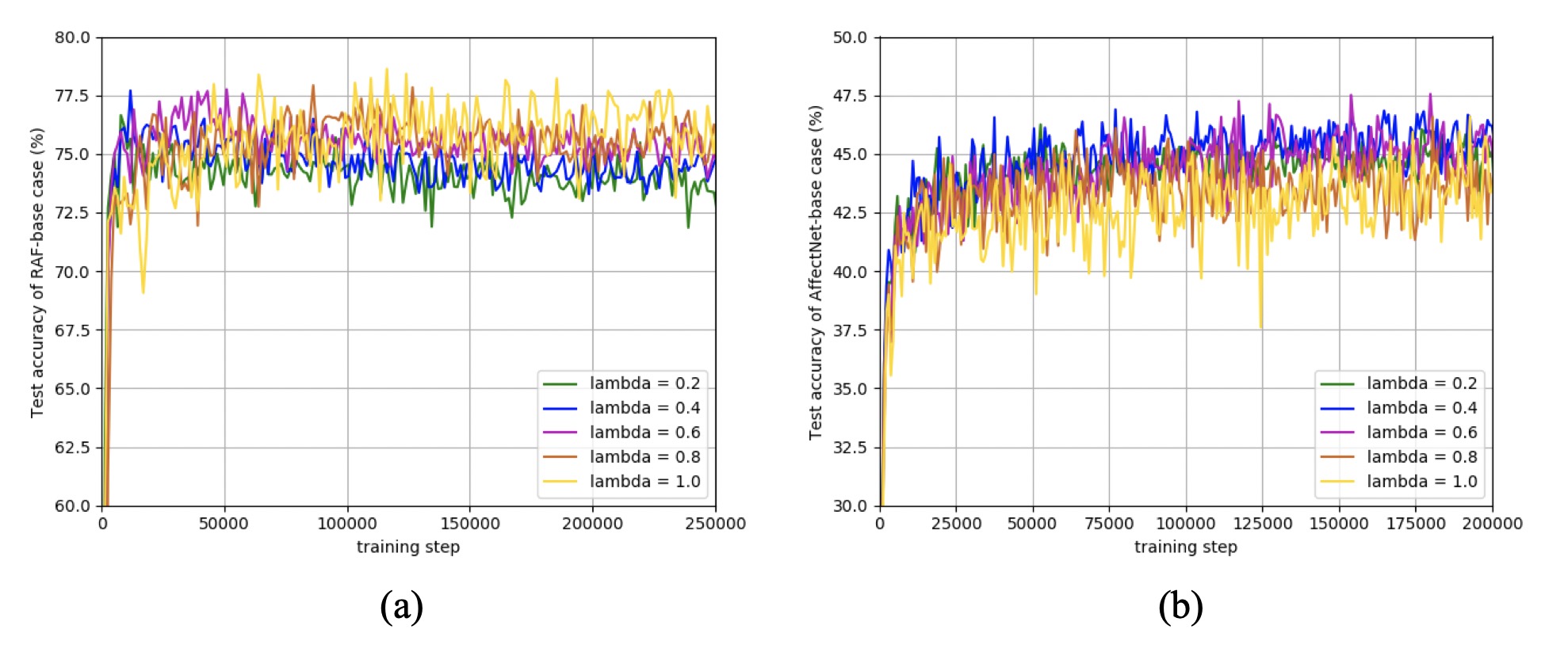}
    \caption{Curve of test accuracy during training with different joint distribution learning weights (i.e. $\lambda$ values) varying among 0.2, 0.4, 0.6, 0.8, 1.0 on (a) RAF-base (b) AffectNet-base cases. The optimal $\lambda$ is selected as 1.0 and 0.4 for RAF-base and AffectNet-base cases respectively.}
    \label{fig:lambda_study}
% \vspace{-0.3cm}
\end{figure}

\begin{figure}[h]
    \centering
    \includegraphics[scale = 0.16]{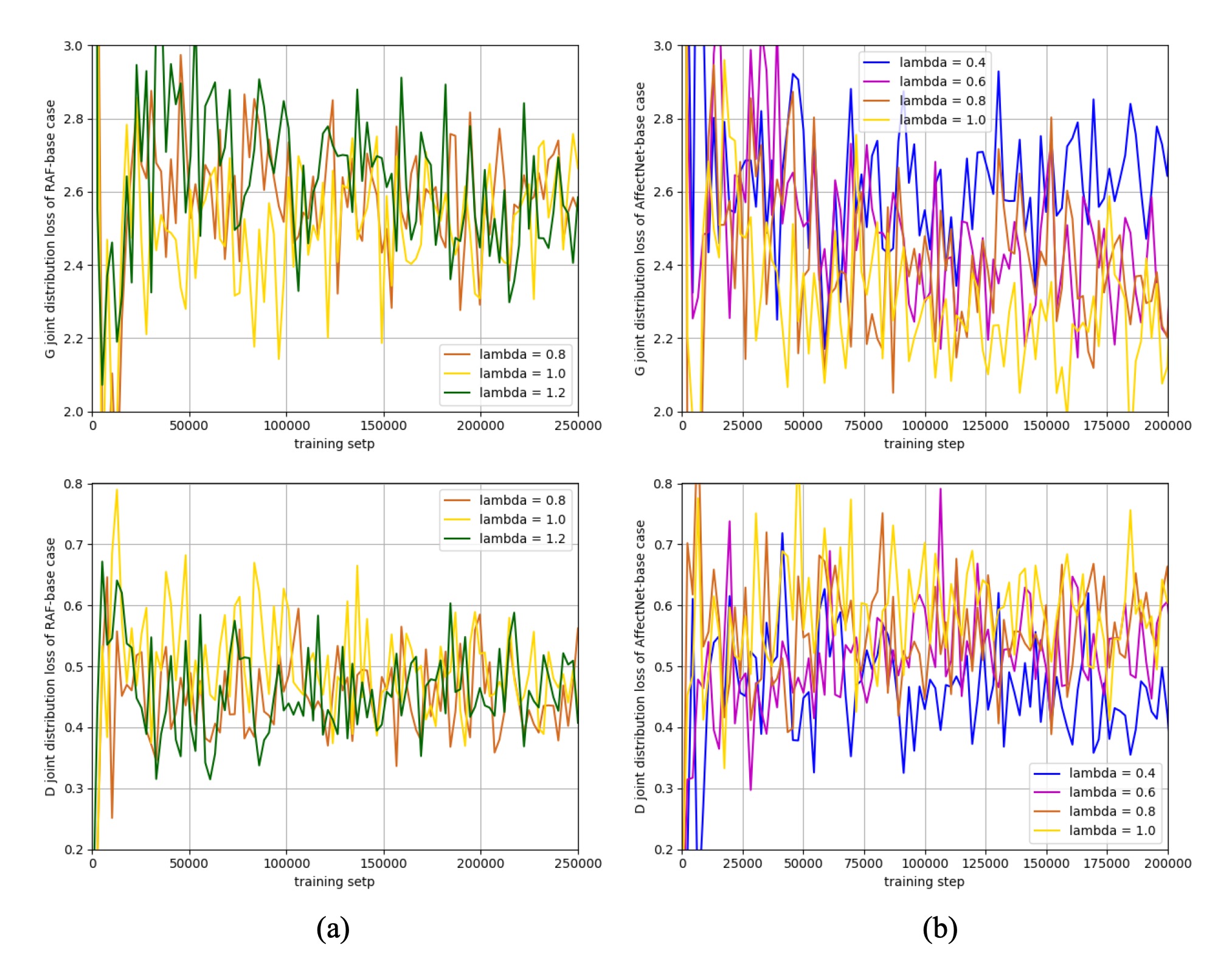}
    \caption{Curve of joint distribution loss during training with different joint distribution learning weights (i.e. $\lambda$ values) (a) RAF-base (b) AffectNet-base cases. The first row is the joint distribution loss of generator (i.e., the encoder $G_Y$ and decoder $G_X$), and the second row is the joint distribution loss of the discriminator $D$).}
    \label{fig:training_loss}
% \vspace{-0.3cm}
\end{figure}

\end{document}